\documentclass[fleqn,twocolumn,10pt]{wlscirep}

\usepackage{graphicx}
\usepackage{float}
\usepackage{amsmath}
\usepackage{amssymb}
\usepackage{multirow}
\usepackage{booktabs}
\usepackage{url}
\usepackage{array}
\usepackage{enumitem}
\usepackage{algorithm}
\usepackage{algorithmic}
\usepackage{pifont}
\usepackage[normalem]{ulem}
\usepackage{makecell}
\usepackage{stackengine}
\usepackage{bm}
\usepackage[misc]{ifsym}
\usepackage{tcolorbox}
\usepackage{subcaption}
\usepackage{tikz}       
\usepackage{fancyhdr}
\newboolean{showlineno}
\setboolean{showlineno}{false} 
\ifthenelse{\boolean{showlineno}}{
  \usepackage[switch]{lineno}
  \linenumbers
}{}

\usepackage{colortbl}
\usepackage{xcolor}

\newcommand{\diffcell}[1]{%
  \ifdim #1pt > 0pt
    \cellcolor{green!#1!white}{#1}
  \else
    \cellcolor{red!-\numexpr#1\relax!white}{#1}
  \fi
}

\usepackage{xparse}
\usepackage{pgfmath}

\def\diffbarW{1.8}   
\def\diffbarH{0.23}  

\NewDocumentCommand{\diffbar}{m g g}{%
  \begin{tikzpicture}[baseline={(barbase.base)}]%
    \IfNoValueTF{#2}{\pgfmathsetmacro{\MIN}{-10}}{\pgfmathsetmacro{\MIN}{#2}}%
    \IfNoValueTF{#3}{\pgfmathsetmacro{\MAX}{10}}{\pgfmathsetmacro{\MAX}{#3}}%

    \pgfmathsetmacro{\W}{\diffbarW}
    \pgfmathsetmacro{\H}{\diffbarH}
    \pgfmathsetmacro{\V}{#1}
    \pgfmathsetmacro{\Z}{(-\MIN)/(\MAX-\MIN)*\W}
    \pgfmathsetmacro{\L}{\V/(\MAX-\MIN)*\W}

    \fill[gray!12] (0,0) rectangle (\W,\H);
    \draw[gray!60, line width=0.2pt] (\Z,0) -- (\Z,\H);

    \ifdim \L cm > 0pt
      \fill[red,opacity=0.4] (\Z,0) rectangle ++(\L,\H);
    \else
      \fill[blue,opacity=0.4] (\Z+\L,0) rectangle (\Z,\H);
    \fi

    \pgfmathsetmacro{\pad}{0.06}
    \coordinate (barbase) at (0,0.22*\H);
    \ifdim \L cm < 0pt
      \node[anchor=west,font=\scriptsize\bfseries]
  		at (\pad,0.5*\H) {\pgfmathprintnumber[fixed,precision=1,zerofill]{\V}\%};      
    \else
  	  \node[anchor=east,font=\scriptsize\bfseries]
  		at (\W-\pad,0.5*\H) {\pgfmathprintnumber[fixed,precision=1,zerofill]{\V}\%};
    \fi
  \end{tikzpicture}%
}

\usepackage[font=small]{caption}
\DeclareCaptionLabelSeparator{pipe}{\space|\space}
\title{Evaluating Large Language Models on Multimodal Chemistry Olympiad Exams}

\author[1,2\thanks{Pre-print version. Check published version at: \url{https://www.nature.com/articles/s42004-025-01782-x}}]{Yiming Cui}
\author[1,2]{Xin Yao}
\author[2]{Yuxuan Qin}
\author[1,2]{Xin Li}
\author[1,2*]{Shijin Wang}
\author[1,2]{Guoping Hu}
\affil[1]{State Key Laboratory of Cognitive Intelligence, Hefei, China}
\affil[2]{iFLYTEK AI Research, Beijing, China}
\affil[*]{\tt ymcui@ieee.org, sjwang3@iflytek.com}

\begin{abstract}\bfseries 
Multimodal scientific reasoning remains a significant challenge for large language models (LLMs), particularly in chemistry, where problem-solving relies on symbolic diagrams, molecular structures, and structured visual data. 
Here, we systematically evaluate 40 proprietary and open-source multimodal LLMs, including GPT-5, o3, Gemini-2.5-Pro, and Qwen2.5-VL, on a curated benchmark of Olympiad-style chemistry questions drawn from over two decades of U.S. National Chemistry Olympiad (USNCO) exams. 
These questions require integrated visual and textual reasoning across diverse modalities. 
We find that many models struggle with modality fusion, where in some cases, removing the image even improves accuracy, indicating misalignment in vision-language integration. 
Chain-of-Thought prompting consistently enhances both accuracy and visual grounding, as demonstrated through ablation studies and occlusion-based interpretability. 
Our results reveal critical limitations in the scientific reasoning abilities of current MLLMs, providing actionable strategies for developing more robust and interpretable multimodal systems in chemistry. 
This work provides a timely benchmark for measuring progress in domain-specific multimodal AI and underscores the need for further advances at the intersection of artificial intelligence and scientific reasoning.
\end{abstract}

\begin{document}

\flushbottom
\maketitle


Multimodal reasoning plays a central role in many scientific domains, where interpreting structured visual information such as diagrams, equations, and charts is essential for problem-solving. Among these, chemistry presents unique challenges due to its reliance on symbolic representations, molecular structures, reaction mechanisms, and experimental data visualizations. As large language models (LLMs) evolve into increasingly capable multimodal LLMs (MLLMs) \cite{hurst2024gpt,bai2025qwen25vl,wang2024qwen2vl,phi4v,dubey2024llama}, their ability to reason in such scientifically structured settings remains poorly understood, especially when compared to recent advances in natural image understanding.

Recent years have witnessed significant progress in visual-language tasks, including image captioning \cite{vinyals2015show}, visual question answering (VQA) \cite{antol2015vqa}, and document understanding \cite{xu2020layoutlm}. Benchmarks like ScienceQA \cite{scienceqa} and MMSci \cite{mmsci} have helped evaluate models in STEM contexts. However, they often focus on physics, math, or general science education. Chemistry, with its highly structured and symbolic multimodal representations, remains underrepresented in this context. OlympiadBench \cite{he-etal-2024-olympiadbench} covers math and physics but omits chemistry, and GPQA \cite{gpqa} assesses graduate-level science knowledge without a visual component.

Chemistry problems are especially challenging for multimodal models due to the abstract and compressed nature of their visual artifacts. Problems may involve Lewis structures, titration curves, electrochemical setups, or molecular conformations, requiring models to understand spatial and symbolic relationships rather than natural scenes. Unlike standard VQA benchmarks, chemistry visuals utilize domain-specific notation and require a deeper integration of visual and textual signals.

Most existing chemistry datasets emphasize perception-level tasks such as chemical structure recognition \cite{rajan2020review,Sadawi2012MolRecAC,brinkhaus2022decimer} or reaction parsing \cite{qian2023rxnscribe}, with relatively fewer efforts directed at integrative and comprehensive reasoning. Some attempts, such as Chemma-RC \cite{chemmarc} and ChemistryQA \cite{wei2021chemistryqa}, extend beyond single-modality recognition toward text-graph integration or multi-hop textual reasoning. However, they remain limited to textual or symbolic inputs and do not incorporate visual modalities. 
To evaluate reasoning in broader contexts, several recent benchmarks have emerged. ChemBench \cite{chembench} provides a large-scale assessment of LLMs’ chemical knowledge, but its scope is restricted to text-only question answering. ChemLLMBench \cite{guo2023can} extends this line of work by incorporating visual modalities, including chemical images and molecular graphs, but primarily focuses on low-level perception tasks such as image-to-graph translation and structure recognition. ChemTable \cite{chemtable} introduces the unique challenge of parsing and interpreting chemical tables, while MaCBench \cite{macbench} offers a comprehensive vision-language evaluation framework spanning data extraction, experimental procedures, and result interpretation in chemistry and materials. Despite these advances, existing resources either remain modality-specific (e.g., text-only, tables, or structure recognition) or emphasize low-level perception and extraction rather than end-to-end multimodal problem solving like that required for Olympiad-style questions.
Parallel to these benchmarks, vision language models such as MolScribe \cite{qian2023molscribe}, RxnIM \cite{rxnim}, ChemDFM-X \cite{chemdfm-x}, ChemVLM \cite{li2025chemvlm}, and ChemMLLM \cite{tan2025chemmllm} have advanced multimodal understanding by linking chemical images, molecular graphs, or spectra to textual representations. These models mark an important step toward multimodal chemical intelligence, yet most are primarily designed for image-to-graph translation, matching tasks, or synthetic cross-modality conversions. They often fall short of supporting open-ended reasoning that integrates diagrams, spectra, and textual problem statements in a holistic, exam-style setting.

\begin{figure*}[ht!]
  \centering
  \includegraphics[width=0.95\textwidth]{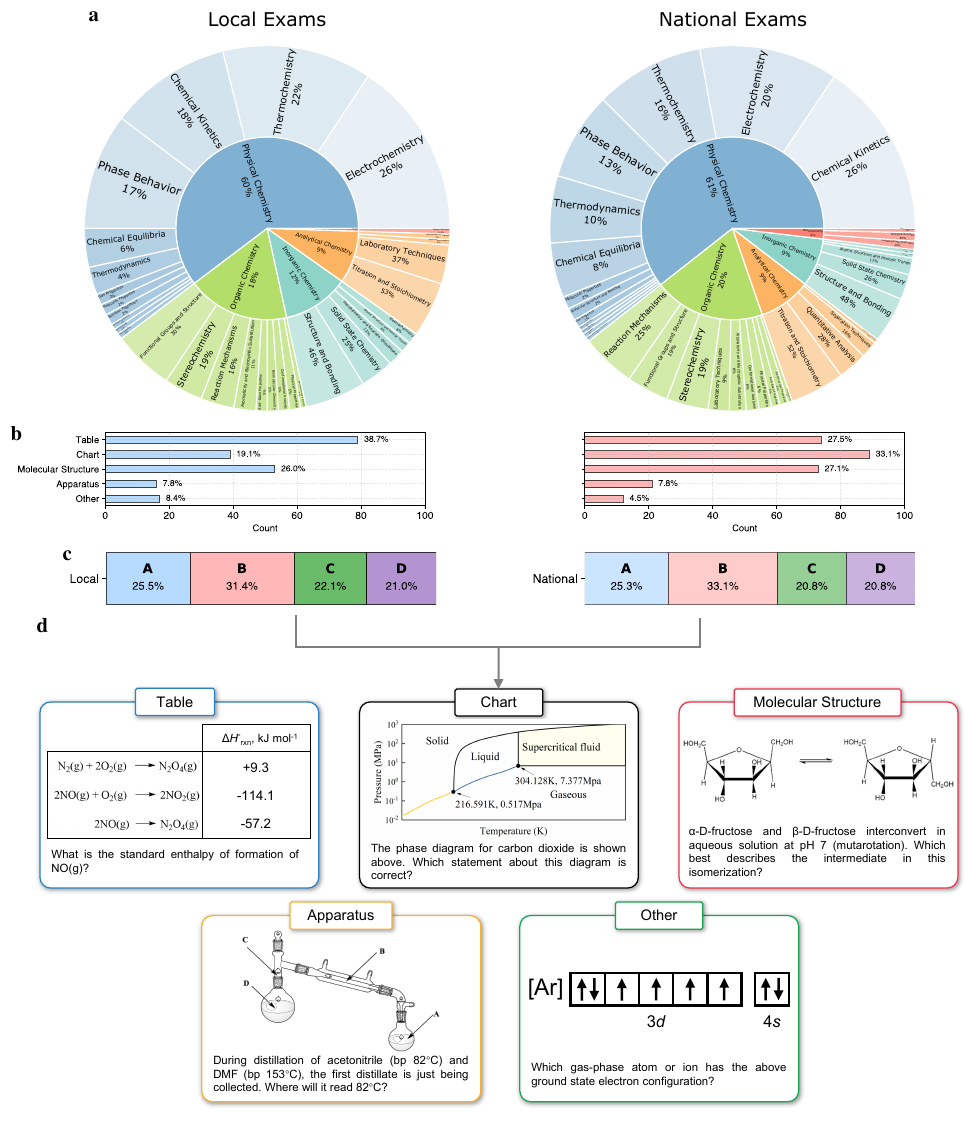}
  \caption{{\bf USNCO-V dataset statistics and example questions.}
  \footnotesize
	{\bf a,} Hierarchy of topic distributions for the local and national sets. The main categories are physical chemistry, organic chemistry, analytical chemistry, inorganic chemistry, and biochemistry.
	{\bf b,} Distributions of image types for local and national sets. Blue bar denotes the local set and red bar denotes the  national set.
	{\bf c,} Distributions of answer choices for the local and national sets.
	{\bf d,} Example questions for different image types (table, chart, molecular structure, apparatus, and other). 
	}
  \label{usnco-v-overview}
\end{figure*}

However, the ability to reason over structured scientific information is critical. Advances such as AlphaFold’s protein structure prediction \cite{jumper2021highly}, deep learning enabled drug discovery \cite{stokes2020deep,frey2023neural}, and automated theorem proving \cite{trinh2024solving} show the transformative potential of scientific AI. These applications rely on models that can engage deeply with symbolic and multimodal data. Chemistry, with its rich visual logic, provides a rigorous testbed for such capabilities.

To address this gap, we curate a chemistry-specific benchmark based on the U.S. National Chemistry Olympiad (USNCO) exams, spanning over two decades (1999--2025). We evaluate 40 multimodal LLMs, including GPT-5, o3, Gemini-2.5-Pro, Qwen2.5-VL, and others, on a curated set of 473 Olympiad-style chemistry questions that require joint visual and textual reasoning. This dataset, USNCO-V, contains diverse visual types (e.g., charts, tables, molecular diagrams, and apparatus) and covers a broad spectrum of chemistry topics, including general, physical, organic, inorganic, and analytical chemistry.

Our analysis includes zero-shot and few-shot prompting experiments, modality ablation, and Chain-of-Thought (CoT) prompting \cite{kojima2022large}. We uncover that small and mid-tier models often suffer from vision-text misalignment, where visual input may degrade performance. While few-shot prompting improves performance for small models, CoT prompting is most effective for mid-scale models, enabling more structured, comparative reasoning. To further understand model behavior, we employ occlusion-based saliency analysis \cite{zeiler2014visualizing,chefer2021transformer}, which reveals how CoT prompting shifts model attention from local pattern matching to global comparison.

These findings suggest that while MLLMs are making strides, their scientific reasoning remains fragile when tasks involve abstract, symbolic, and visually grounded content. The metadata of USNCO-V is released publicly to enable reproducibility and facilitate future benchmarking. By providing a targeted and interpretable evaluation of MLLMs in chemistry, this work offers actionable insights into multimodal fusion, prompting strategies, and model diagnostics for the next generation of scientific AI.

\section*{Results}

\subsection*{Overview of USNCO-V}

We curated a chemistry-specific multimodal benchmark using historical U.S. National Chemistry Olympiad exams, which are publicly available in PDF format, spanning local exams (2000--2025) and national exams (1999--2025). 
Each local exam consists of 60 multiple-choice questions. In comparison, the national exams comprise three parts: Part One (60 multiple-choice questions), Part Two (open-ended problem-solving), and Part Three (laboratory-based tasks). 
This study focuses exclusively on the multiple-choice portions from both local and national exams, which can be accurately evaluated based on ground truth answers.

The resulting dataset, namely USNCO-V (where `V' stands for vision), consists of 473 multimodal questions. Each problem consists of a textual question, an image (e.g., a diagram, chart, molecular structure, or experimental apparatus), and four answer choices, covering the categories table, chart, molecular structure, apparatus, and other. 
The USNCO-V mainly covers physical, organic, analytical, and inorganic chemistry, as well as a small portion of biochemistry (Fig.~\ref{usnco-v-overview}a).
The distribution of image types is depicted in Fig.~\ref{usnco-v-overview}b.
The distribution of correct options is balanced across options, with slightly more option `B' in both sets, reducing bias risk (Fig.~\ref{usnco-v-overview}c).
A set of example questions for different image types is shown in Fig.~\ref{usnco-v-overview}d.
A detailed dataset curation pipeline is described in the Methods section.

\begin{table*}[!t]
\footnotesize
\begin{center}
\caption{\label{results-proprietary-models} {\bf Results of proprietary models on USNCO-V.} \footnotesize 
	Acc (95\% CI) indicates accuracy with its 95\% confidence interval. Pass@5 measures the percentage of questions answered correctly in at least one of five attempts. Cons@5 (consistency@5) represents consistency across five responses. Average is the mean of local and national (macro-)accuracies, and Difference shows the performance gap between the two subsets, with negative values favoring the national set. }
\begin{tabular}{l cccc cccc c c }
\toprule
\multirow{2}{*}{\bf Model} & \multicolumn{3}{c}{\bf Local Set} & \multicolumn{3}{c}{\bf National Set} & \multirow{2}{*}{\bf Average} & \multirow{2}{*}{\bf Difference}  \\ 
& {\bf Acc (95\% CI)} & {\bf Pass@5} & {\bf Cons@5} & {\bf Acc (95\% CI)} & {\bf Pass@5} & {\bf Cons@5}  \\
\midrule
GPT-5 				& 93.7 \tiny{[90.7-96.6]} & 96.1 & 97.6 & 92.7 \tiny{[90.0-95.2]} & 97.4 & 96.6 & 93.2 & \diffbar{1.00}{-5}{5} \\
Gemini-2.5-Pro 		& 91.4 \tiny{[87.7-94.7]} & 94.6 & 97.4 & 90.6 \tiny{[87.7-93.3]} & 95.5 & 95.2 & 91.0 & \diffbar{0.80}{-5}{5} \\
o3 					& 91.2 \tiny{[87.5-94.6]} & 94.6 & 96.1 & 89.2 \tiny{[85.9-92.6]} & 92.6 & 95.9 & 90.2 & \diffbar{2.00}{-5}{5} \\
Gemini-2.5-Flash 	& 89.6 \tiny{[85.9-92.8]} & 96.6 & 94.8 & 87.9 \tiny{[84.8-90.9]} & 96.7 & 93.0 & 88.8 & \diffbar{1.70}{-5}{5} \\
o4-mini 			& 87.7 \tiny{[83.8-91.4]} & 95.6 & 94.6 & 87.7 \tiny{[84.5-90.6]} & 96.7 & 93.0 & 87.7 & \diffbar{0.09}{-5}{5} \\
GPT-5-mini 			& 88.2 \tiny{[84.4-91.9]} & 94.1 & 94.2 & 86.8 \tiny{[83.4-90.1]} & 94.4 & 94.6 & 87.5 & \diffbar{1.40}{-5}{5} \\
o1 					& 86.2 \tiny{[81.8-90.4]} & 92.2 & 95.4 & 84.1 \tiny{[80.4-87.8]} & 92.9 & 93.7 & 85.2 & \diffbar{2.10}{-5}{5} \\
GPT-5-nano 			& 77.9 \tiny{[73.1-82.5]} & 89.7 & 89.3 & 78.2 \tiny{[74.1-82.2]} & 91.1 & 89.4 & 78.1 & \diffbar{-0.26}{-5}{5} \\
GPT-4.1 			& 57.6 \tiny{[51.4-63.8]} & 67.2 & 92.5 & 56.7 \tiny{[51.3-62.1]} & 67.7 & 91.8 & 57.2 & \diffbar{0.90}{-5}{5} \\
GPT-4.1-mini 		& 48.4 \tiny{[42.5-54.4]} & 64.2 & 85.8 & 51.6 \tiny{[46.3-56.8]} & 65.1 & 86.8 & 50.0 & \diffbar{-3.20}{-5}{5} \\
GPT-4o 				& 46.7 \tiny{[41.0-52.5]} & 67.2 & 82.9 & 50.4 \tiny{[45.5-55.4]} & 72.9 & 79.4 & 48.6 & \diffbar{-3.70}{-5}{5} \\
GPT-4.1-nano 		& 32.3 \tiny{[27.0-37.6]} & 48.5 & 82.3 & 31.9 \tiny{[27.2-36.4]} & 49.1 & 81.4 & 32.1 & \diffbar{0.40}{-5}{5} \\
\bottomrule
\end{tabular}
\end{center}
\end{table*}

\begin{table*}[!t]
\footnotesize
\begin{center}
\caption{\label{results-oss-models} {\bf Results of open-source models on USNCO-V.}}
\begin{tabular}{l c c c c}
\toprule
{\bf Model} & {\bf Local Acc (95\% CI)} & {\bf National Acc (95\% CI)} & {\bf Average} & {\bf Difference}  \\ 
\midrule
\multicolumn{4}{l}{\em Open-source Models (large-level: $\ge$ 20B)} \\
Intern-S1 (241B)				& 59.8 \tiny{[53.0-66.3]} & 56.5 \tiny{[50.5-62.3]} & 58.2 & \diffbar{3.3} \\
InternVL3.5-38B				& 55.9 \tiny{[49.0-62.5]} & 52.4 \tiny{[46.5-58.3]} & 54.2 & \diffbar{3.5} \\ 
InternVL3-78B				& 57.8 \tiny{[51.0-64.4]} & 49.1 \tiny{[43.2-55.0]} & 53.5 & \diffbar{8.7} \\ 
Qwen2.5-VL-72B-Instruct	 	& 52.0 \tiny{[45.1-58.7]} & 51.7 \tiny{[45.7-57.6]} & 51.8 & \diffbar{0.3} \\
InternVL3.5-30B-A3B			& 51.5 \tiny{[44.6-58.2]} & 47.2 \tiny{[41.3-53.2]} & 49.3 & \diffbar{4.3} \\
Qwen2-VL-72B-Instruct	 	& 49.0 \tiny{[42.2-55.8]} & 48.0 \tiny{[42.1-53.9]} & 48.5 & \diffbar{1.1} \\
Qwen2.5-VL-32B-Instruct	 	& 50.5 \tiny{[43.7-57.3]} & 46.1 \tiny{[40.2-52.1]} & 48.3 & \diffbar{4.4} \\
InternVL3.5-20B-A4B			& 40.2 \tiny{[33.7-47.0]} & 35.3 \tiny{[29.8-41.2]} & 37.8 & \diffbar{4.9} \\
Gemma-3-27B-IT			 	& 35.8 \tiny{[29.5-42.6]} & 33.1 \tiny{[27.7-38.9]} & 34.4 & \diffbar{2.7} \\
Molmo-72B-0924				& 31.9 \tiny{[25.9-38.5]} & 35.7 \tiny{[30.2-41.6]} & 33.8 & \diffbar{-3.8} \\ 
\midrule
\multicolumn{4}{l}{\em Open-source Models (base-level: $<$20B)} \\
InternVL3.5-14B				& 47.1 \tiny{[40.3-53.9]} & 48.0 \tiny{[42.1-53.9]} & 47.5 & \diffbar{-0.9} \\
Ovis2.5-9B					& 52.0 \tiny{[45.1-58.7]} & 42.4 \tiny{[36.6-48.4]} & 47.2 & \diffbar{9.6} \\
Intern-S1-Mini (8B) 			& 49.0 \tiny{[42.2-55.8]} & 40.9 \tiny{[35.2-46.9]} & 45.0 & \diffbar{8.1} \\
InternVL3.5-4B 				& 39.7 \tiny{[33.2-46.6]} & 47.6 \tiny{[41.7-53.5]} & 43.7 & \diffbar{-7.9} \\
InternVL3.5-8B 				& 40.7 \tiny{[34.2-47.5]} & 45.4 \tiny{[39.5-51.3]} & 43.0 & \diffbar{-4.7} \\
GLM-4.1V-9B-Thinking			& 43.6 \tiny{[37.0-50.5]} & 41.3 \tiny{[35.5-47.2]} & 42.5 & \diffbar{2.4} \\
MiniCPM-V 4.5				& 39.7 \tiny{[33.2-46.6]} & 45.4 \tiny{[39.5-51.3]} & 42.5 & \diffbar{-5.6} \\
Qwen2.5-VL-7B-Instruct	 	& 36.8 \tiny{[30.5-43.6]} & 39.0 \tiny{[33.4-45.0]} & 37.9 & \diffbar{-2.3} \\
Ovis2.5-2B					& 37.3 \tiny{[30.9-44.1]} & 28.6 \tiny{[23.6-34.3]} & 32.9 & \diffbar{8.6} \\
Phi-4-Multimodal (5.6B)		& 33.3 \tiny{[27.2-40.1]} & 31.6 \tiny{[26.3-37.4]} & 32.5 & \diffbar{1.7} \\
Gemma-3-12B-IT				& 36.3 \tiny{[30.0-43.1]} & 26.8 \tiny{[21.8-32.4]} & 31.5 & \diffbar{9.5} \\
InternVL3.5-2B 				& 29.9 \tiny{[24.0-36.5]} & 31.6 \tiny{[26.3-37.4]} & 30.8 & \diffbar{-1.7} \\
Qwen2.5-VL-3B-Instruct	 	& 30.9 \tiny{[24.9-37.5]} & 30.5 \tiny{[25.3-36.2]} & 30.7 & \diffbar{0.4} \\
Molmo-7B-D-0924				& 32.8 \tiny{[26.8-39.6]} & 26.0 \tiny{[21.1-31.6]} & 29.4 & \diffbar{6.8} \\
Gemma-3-4B-IT				& 25.5 \tiny{[20.0-31.9]} & 26.8 \tiny{[21.8-32.4]} & 26.1 & \diffbar{-1.3} \\
Janus-Pro-7B					& 24.5 \tiny{[19.1-30.8]} & 26.4 \tiny{[21.5-32.0]} & 25.4 & \diffbar{-1.9} \\
\midrule
\multicolumn{4}{l}{\em Chemistry-specific Models} \\
ChemVLM-8B					& 30.9 \tiny{[24.9-37.5]} & 29.4 \tiny{[24.2-35.1]} & 30.1 & \diffbar{1.5} \\
ChemVLM-26B					& 31.9 \tiny{[25.9-38.5]} & 26.0 \tiny{[21.1-31.6]} & 28.9 & \diffbar{5.8} \\
\bottomrule
\end{tabular}
\end{center}
\end{table*}

\subsection*{Baseline performance on multimodal chemistry problems}

We evaluate 40 state-of-the-art multimodal large language models, including both proprietary systems (e.g., GPT-5\cite{gpt-5}, Gemini-2.5-Pro\cite{google2025gemini25pro}) and open-source models (e.g., Qwen2.5-VL\cite{bai2025qwen25vl},  InternVL3.5\cite{internvl35}, Molmo\cite{Deitke2024MolmoAP}, Phi-4-Multimodal\cite{phi4v}) on USNCO-V.
For baseline evaluation, we use the zero-shot prompting method, which is widely used in benchmarking LLMs' performance.
Detailed experimental setups and model lists are available in the Methods section.

Table \ref{results-proprietary-models} shows a clear top tier among proprietary MLLMs on USNCO-V. GPT-5 leads with a 93.2\% macro-average, followed closely by Gemini-2.5-Pro (91.0\%) and o3 (90.2\%). Gemini-2.5-Flash, o4-mini, and GPT-5-mini follow closely in the high-80s, while o1 and GPT-5-nano form the next tier, ranging from the mid-80s down to the mid-70s. Models explicitly optimized for reasoning, such as the GPT-5 series, o-series, and Gemini-2.5 family, consistently outperform non-reasoning models like GPT-4.1 and GPT-4o, highlighting the importance of advanced reasoning alignment for chemistry problem solving.
For all proprietary models, pass@5 provides additional gains over accuracy, showing only modest improvements for top-performing systems but increases of up to 20 percentage points (pp) for weaker ones (e.g., GPT-4o), implying that many errors can be recovered through stochastic decoding. Meanwhile, consistency@5 remains high (often >90\%) even when accuracy is moderate, as seen with GPT-4.1, pointing to confident but systematic failure modes. Smaller variants, such as GPT-5-nano, perform about 15 pp lower than GPT-5, underscoring how both model scale and specialized training are critical for achieving exam-level multimodal reasoning.

Table~\ref{results-oss-models} presents the results for open-source MLLMs. At the large-level ($\geq$20B parameters), performance peaks with Intern-S1 (241B) at 58.2\%, followed by InternVL3.5-38B (54.2\%) and InternVL-3-78B (53.5\%). Other 30--70B models, such as Qwen2.5-VL-72B-Instruct (51.8\%) and InternVL3.5-30B-A3B (49.3\%), achieve near-50s accuracy, while models like Gemma-3-27B-IT and Molmo-72B fall below 35\%. These results underscore that scaling alone does not guarantee improved performance, while instruction tuning and dataset quality play a crucial role in determining outcomes. 	

Smaller open-source models (<20B) show wide variability. Some mid-range systems, such as InternVL3.5-14B (47.5\%), Ovis2.5-9B (47.2\%), and Intern-S1-Mini (8B) (45.0\%), approach the performance of much larger counterparts, whereas others, including Janus-Pro-7B (25.4\%) and Gemma-3-4B-IT (26.1\%), remain far behind. Interestingly, several compact models outperform larger ones (e.g., Qwen2.5-VL-7B at 37.9\% surpasses Molmo-72B at 33.8\%), indicating that knowledge coverage and alignment strategies can partially offset a smaller scale.

Despite these advances, open-source models remain 30--35 pp behind the top proprietary systems. Even the strongest open-source reasoning models, such as Intern-S1 and Qwen2.5-VL-72B-Instruct, lag substantially, showing that progress in scientific multimodal reasoning requires not just scale, but also richer chemistry-aligned training. Overall, open-source models display encouraging improvements in general multimodal capability, yet continue to trail proprietary models in both reasoning depth and exam-level domain knowledge.
Although many open-source MLLMs report results comparable to proprietary models on broad multimodal benchmarks, our findings show that when evaluated on a chemistry-specialized benchmark such as USNCO-V, the limits of domain knowledge coverage become apparent. This underscores the importance of discipline-specific evaluation for assessing scientific reasoning capabilities beyond generic multimodal understanding.

Across both proprietary and open-source models, most perform slightly better on the local subset than on the national subset, with typical margins of +1--3 pp. These differences are minimal among top-tier models (e.g., GPT-5 and Gemini-2.5-Pro, with $\le$+1 pp), but a few lower-tier models invert this trend, such as GPT-4o (-3.7) and GPT-4.1-mini (-3.2), suggesting sensitivity to subset-specific content and distributional differences. Importantly, our ablation experiments show that removing the image modality causes a larger performance drop on the national set, indicating that national exams require stronger multimodal integration. This suggests that subset-level differences arise not only from textual difficulty but also from the degree to which visual reasoning is needed, an issue we examine in detail in the next section.

In addition to general-purpose MLLMs, we also evaluated two chemistry-specialized MLLMs, ChemVLM-8B and ChemVLM-26B. Both models achieve modest accuracies of 30.1\% and 28.9\%, respectively, which are comparable to small open-source baselines but substantially lower than state-of-the-art proprietary systems. 
Notably, the larger ChemVLM variant does not outperform its smaller counterpart, suggesting that scaling within a perception-focused training regime does not directly translate to stronger performance on exam-style reasoning tasks. 
These results highlight a critical distinction: while chemistry MLLMs excel at structure recognition and graph translation, they struggle to generalize to integrative multimodal reasoning required in Olympiad-style assessments. 
This finding underscores the importance of benchmarks like USNCO-V for probing scientific reasoning capabilities beyond perception, and demonstrates that domain specialization alone is insufficient without explicit training for reasoning across textual and visual modalities.

\subsection*{Few-Shot Prompting Helps Smaller Models, but Not Always}

\begin{figure*}[ht!]
  \centering
  \includegraphics[width=1\textwidth]{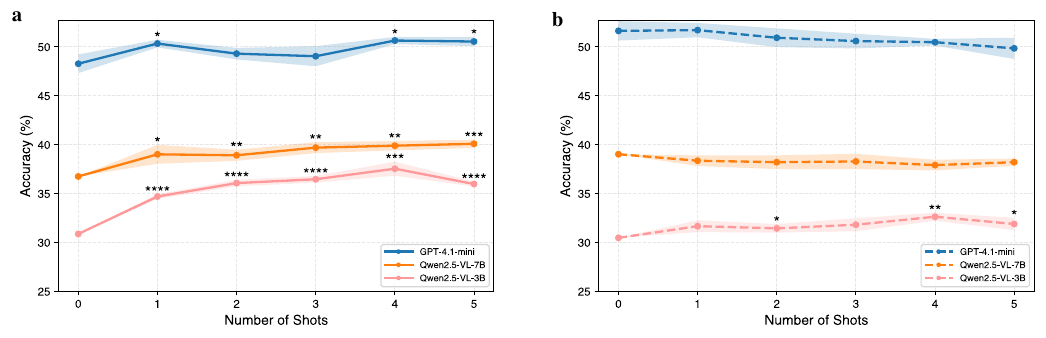}
  \caption{{\bf Results of few-shot prompting.} \footnotesize Performance of Qwen2.5-VL-3B, Qwen2.5-VL-7B, and GPT-4.1-mini on the local and national subsets using 0-shot to 5-shot prompting.
  	{\bf a,} Performance on the local set.
  	{\bf b,} Performance on the national set.
  	Zero-shot corresponds to no example provided, while higher shot counts include more in-context examples. Lines indicate mean accuracy, and shaded ribbons represent the standard error of the mean across five runs. Statistical significance is indicated only for improvements over the zero-shot baseline, based on one-sided Welch's $t$-tests comparing accuracy across five runs.
  	$^{*}p{<}0.05$; $^{**}p{<}0.01$; $^{***}p{<}0.001$; $^{****}p{<}0.0001$.
  }
  \label{fewshot-results}
\end{figure*}

We evaluate the effectiveness of few-shot prompting \cite{brown2020language} on three models: Qwen2.5-VL-3B, Qwen2.5-VL-7B, and GPT-4.1-mini. Prompts are formatted as multi-turn dialogues using examples sampled from the opposite subset (e.g., national examples for local evaluation and vice versa). Fig~\ref{fewshot-results} shows accuracy trends from zero-shot to five-shot settings, with statistical significance indicated by one-sided Welch's $t$-tests relative to the zero-shot baseline.

Qwen2.5-VL-3B exhibits the most significant relative improvement. On the local set, accuracy rises steadily from 30.9\% at zero-shot to 37.5\% at 4-shot ($p{<}0.001$), with statistically significant gains appearing as early as the 1-shot condition. Performance on the national set also improves but more modestly, peaking at 32.6\% around 4-shot ($p{<}0.01$). Notably, significance markers are denser on the local set, indicating that few-shot prompting provides more reliable improvements when tasks are simpler and primarily text-driven.

Qwen2.5-VL-7B shows moderate and steady gains on the local set, rising from 36.8\% at zero-shot to 40.1\% at 5-shot, with only minor fluctuations. In contrast, performance on the national set slightly declines after zero-shot, suggesting that added examples do not provide clear benefits and may even introduce noise. These patterns indicate that while mid-scale models benefit from few-shot prompting on simpler, text-driven tasks, they can quickly saturate or destabilize when tasks demand higher multimodal reasoning.

GPT-4.1-mini, despite being a proprietary model, shows only marginal changes, rising slightly from 48.3\% to 50.7\% on the local set and declining from 51.6\% to 49.9\% on the national set as shot counts increase. This stability suggests that its zero-shot reasoning is already well-calibrated for multimodal chemistry problems, and additional examples can sometimes introduce noise rather than clarity.

Taken together, these results reveal a clear contrast between subsets. Few-shot prompting is highly effective on the local set, where questions are easier and more text-centric, but less impactful on the national set, which contains more complex, visually intensive problems. This aligns with our image ablation findings (in the following section), where removing images caused larger performance drops on the national set. These converging lines of evidence indicate that national exams require deeper multimodal integration, which cannot be overcome simply by adding more textual demonstrations. Thus, while few-shot prompting is a valuable strategy for smaller models and text-heavy tasks, tackling the most challenging visual reasoning questions will require architectural advances rather than prompting alone.

\subsection*{CoT Prompting: Helpful for All, Crucial for Mid-Tier Models}

\begin{figure}[t]
  \centering
  \includegraphics[width=1\columnwidth]{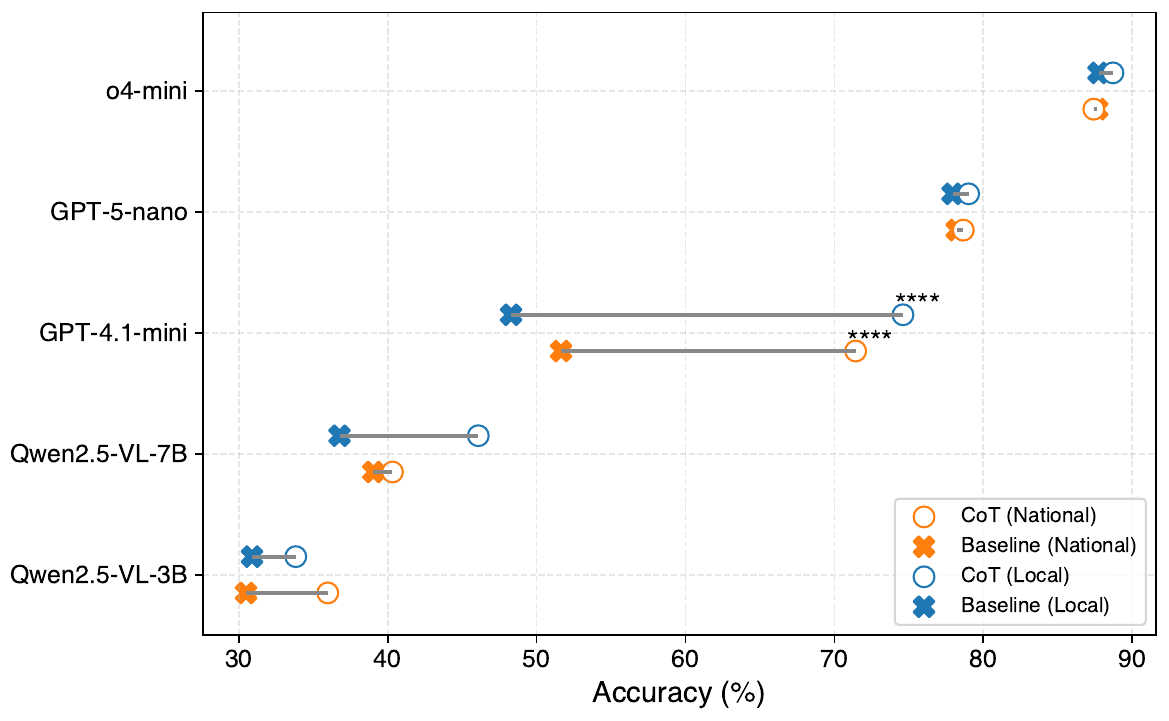}
  \caption{\label{cot-results} {\bf Results of Chain-of-Thought prompting.}
  \footnotesize
Dumbbell plot of baseline ({\bf x}) and CoT ({\bf o}) accuracies for each model on the local (top, blue) and national (bottom, orange) USNCO-V sets. 
Lines connect baseline $\rightarrow$ CoT for each model, using mean accuracy across five runs. 
Statistical significance is indicated only for models with repeated trials (GPT-4.1-mini, GPT-5-nano, o4-mini) based on one-sided Welch's $t$-tests.
Significance markers are omitted for Qwen2.5-VL-3B and 7B, as their results are deterministic and lack measurable variance.
$^{*}p{<}0.05$; $^{**}p{<}0.01$; $^{***}p{<}0.001$; $^{****}p{<}0.0001$.}
\end{figure}

To assess the impact of explicit reasoning strategies, we applied Chain-of-Thought (CoT) prompting \cite{kojima2022large} to several evaluated MLLMs using a standardized template that encourages step-by-step problem-solving. Fig.~\ref{cot-results} presents dumbbell plots comparing baseline and CoT prompting accuracies across the local and national subsets of USNCO-V.

Among all models, GPT-4.1-mini exhibits the most substantial improvement, gaining 26.3 pp on the local subset (48.3→74.6, $p{<}0.0001$) and nearly 20 pp on the national subset (51.6→71.4, $p{<}0.0001$). This dramatic increase underscores how mid-capacity models, those with adequate reasoning potential but limited implicit structure, benefit most from CoT scaffolding. Qwen2.5-VL-7B also sees meaningful gains, particularly on the local subset (+9.3 pp; 36.8→46.1), though its improvement on the national subset is smaller (+1.3 pp; 39.0→40.3), reinforcing the trend that mid-tier models are especially responsive to structured prompting in visually grounded tasks.

In contrast, Qwen2.5-VL-3B improves more modestly: +2.9 pp locally (30.9→33.8) and +5.5 pp nationally (30.5→36.0). These smaller jumps suggest that while CoT may aid lower-capacity models, its utility is bounded by the model’s ability to generate or follow structured reasoning chains. Meanwhile, top-performing proprietary models such as o4-mini and GPT-5-nano experience only marginal gains ($\le$1.1 pp), reflecting a clear ceiling effect. These powerful reasoning models already contain implicit reasoning internally, limiting the marginal benefit of additional prompt structure.

Interestingly, this outcome contrasts with our few-shot prompting results (see Fig.~\ref{fewshot-results}), where small models such as Qwen2.5-VL-3B benefited the most. This divergence suggests that while small models can imitate demonstrations in few-shot settings, they struggle to self-generate reasoning chains in CoT prompting, which demands autonomous logical construction. In contrast, mid-tier models like GPT-4.1-mini excel when scaffolded, improving by over 26 pp on the local subset, highlighting their latent capacity for structured reasoning. Together, these results reveal a U-shaped response curve: small models lack reasoning capacity, large models already possess it, and mid-tier models benefit the most from CoT-style guidance.

These findings point to a promising direction for future model development. CoT prompting acts as an external reasoning scaffold that bridges the gap between heuristic pattern-matching and systematic logic. For mid-scale models, it can unlock substantial latent capacity; for large models, it offers marginal but reliable gains; and for small models, effectiveness may depend on further simplification or architectural refinements. Future work may focus on embedding CoT-like structures into pre-training objectives or fine-tuning paradigms to foster native stepwise reasoning, particularly in scientific domains that demand interpretability and rigor.

\subsection*{Task-Type Breakdown Reveals Model Specialization and Weaknesses}

To investigate how multimodal models perform across different image types, we analyze five visual task categories: tables, charts, molecular structures, apparatus, and other. Accuracy is reported separately for flagship proprietary systems (GPT-5, Gemini-2.5-Pro) and for mid-tier (GPT-4.1, GPT-4o) and open-source baselines (Qwen2.5-VL-7B, Qwen2.5-VL-3B).

\begin{figure}[t!]
  \centering
  \includegraphics[width=1\columnwidth]{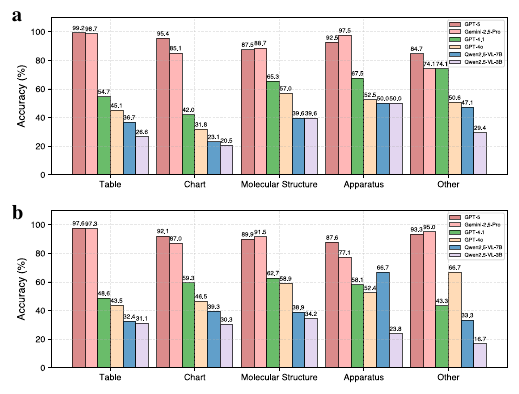}
  \caption{{\bf Per-category performance on USNCO-V.} \footnotesize Each question is classified into one of five visual categories: table, chart, molecular structure, apparatus, or other. {\bf a,} Accuracy breakdown for the local set. {\bf b,} Accuracy breakdown for the national set.}
  \label{per-task-results}
\end{figure}

As shown in Fig.~\ref{per-task-results}, GPT-5 and Gemini-2.5-Pro achieve near-ceiling performance on tables ($\ge$97\%) and consistently strong results on charts ($\ge$85\%), reflecting their pre-training on abundant generic visual formats. By contrast, their accuracy on chemistry-specific categories is slightly lower: for GPT-5, molecular structures and apparatus remain in the high-80s to low-90s, while Gemini-2.5-Pro is very strong on local apparatus but drops on national apparatus (from 97.5\% to 77.1\%). This gap suggests that while these models have mastered general-purpose visual reasoning, domain-specific encodings such as bond-line diagrams or experimental schematics remain more challenging. The distinction highlights where chemistry-specific training data would most effectively improve future systems.

Mid-tier proprietary models show mixed patterns across categories. GPT-4.1 performs strongest on molecular structures (about 63--65\%) and apparatus (about 58--68\%), while lagging on tables (about 49--55\%) and achieving moderate scores on charts (about 42--59\%). GPT-4o is less consistent, performing poorly on charts (about 32--47\%) but reaching moderate accuracy on molecular structures and apparatus (about 57--59\% and 52--53\%, respectively). Together, these results suggest that mid-tier models can handle structured chemical diagrams reasonably well but remain weaker on tabular data and quantitative plots, pointing to partial rather than full multimodal integration.

\begin{figure*}[t!]
  \centering
  \includegraphics[width=0.95\textwidth]{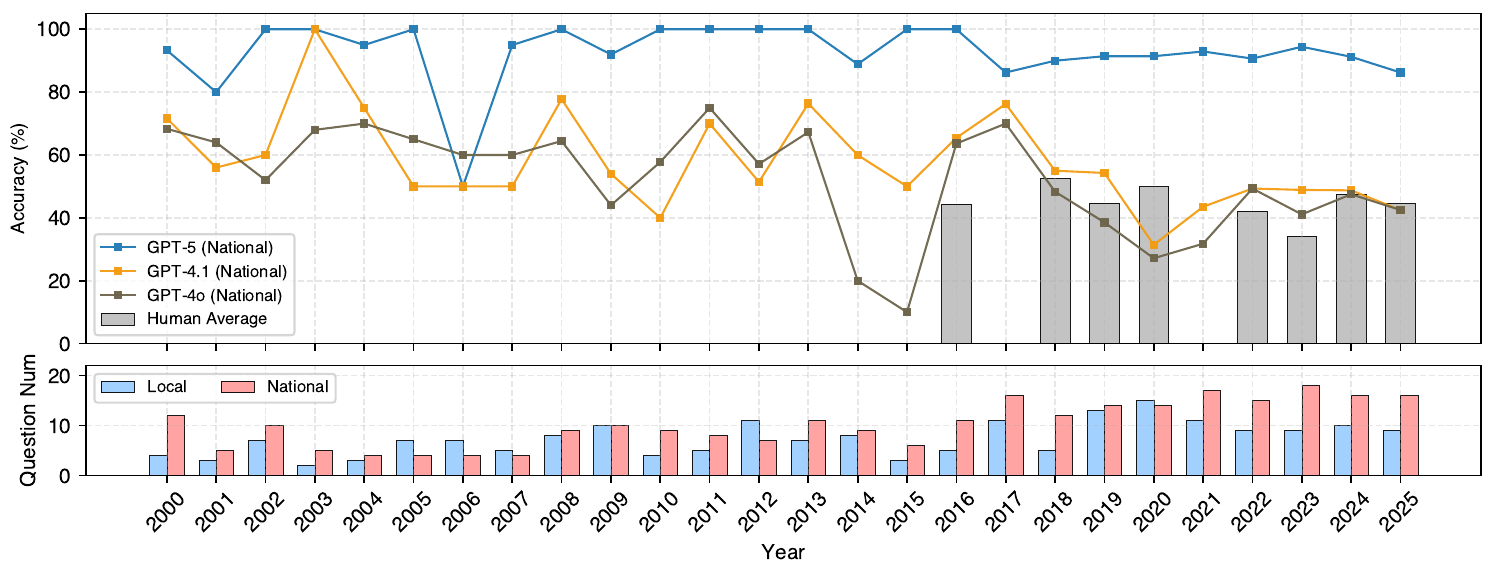}
  \caption{\label{human-results} {\bf Top-performing model and student participant performance over time on USNCO-V.} 
  \footnotesize
  \textbf{Top:} Accuracy of GPT-5, GPT-4.1, and GPT-4o on the national subset, compared with the estimated scores of human participants (red bars). \textbf{Bottom:} Number of multimodal questions in each year's test set for the local and national exams.}  
\end{figure*}

Smaller open-source models remain substantially weaker overall, though some categories reveal relative strengths. Qwen2.5-VL-7B performs best on apparatus tasks (50\% local, 67\% national), while Qwen2.5-VL-3B achieves modest gains on molecular structures (34--40\%). Both models, however, struggle most on charts (20--39\%), underscoring chart-based reasoning as a persistent challenge at smaller scales. These patterns suggest that weaker models are more easily misled by cluttered or quantitative visuals, whereas schematic apparatus depictions may offer clearer inductive cues.

Cross-subset differences further illuminate model behavior. For non-top models, chart accuracy is consistently higher on the national subset, whereas apparatus questions often show the opposite trend (with several models performing worse on the national set). Notable exceptions include Qwen2.5-VL-7B, which improves on national apparatus, and GPT-4o, which is nearly flat across subsets. The heterogeneous `other' category exhibits high variance across models and subsets but is less interpretable. We therefore focus our analysis on the four more chemically meaningful categories.

Taken together, these results reveal that models already excel at tasks requiring generic visual literacy (tables, charts) but remain less reliable on chemistry-specific modalities (molecular structures, apparatus) that demand domain knowledge and visual intuition. For chemists, this distinction underscores both the promise and the current limits of multimodal LLMs: they can already assist with data interpretation but are not yet dependable for structural recognition or experimental reasoning. This gap highlights the value of benchmarks like USNCO-V in pinpointing where chemistry-specific visual skills remain a frontier for AI.

\subsection*{Surpassing Human Average: Model Accuracy by Year}

To contextualize model performance relative to human benchmarks, we compare the accuracy of three representative GPT models (GPT-5, GPT-4.1, and GPT-4o) against estimated scores of USNCO participants. 
Fig.~\ref{human-results} shows the annual accuracy for the national subset, along with human benchmark scores derived from publicly reported cutoff data where available (years: 2025, 2024, 2023, 2022, 2020, 2019, 2018, and 2016).

Across all overlapping years, GPT-5 consistently and substantially outperforms the human benchmark on the national subset. 
In 2025, it attains 86.3\% (13.8/16) on the national set versus an estimated human score of 44.6\%. This performance gap persists in earlier years (e.g., +41.5 pp in 2020 and +60.2 pp in 2023), demonstrating the growing competence of MLLMs in complex, visually grounded chemistry tasks. Importantly, GPT-5 remains strong even on earlier exams (e.g., 2010--2016), which include relatively few multimodal questions.

For mid-tier systems, GPT-4.1 generally exceeds the human benchmark on the national subset in most overlapping years (2016, 2018, 2019, 2022--2024), while trailing in 2020 and 2025. For example, in 2020, GPT-4.1 reaches 31.4\% (4.4/14) versus an estimated human score of 49.9\%, and in 2025, it attains 42.5\% (6.8/16) versus 44.6\%. GPT-4o exhibits greater variability: it exceeds the human benchmark in 2016, 2022, and 2023, is near parity in 2024, and falls below in the remaining overlapping years, with a pronounced gap in 2020 (27.1\% or 3.8/14 versus 49.9\%). These comparisons indicate that while both GPT-4.1 and GPT-4o often surpass the estimated human average, their performance is less uniform across years than GPT-5.

The bottom panel reveals a marked upward trend in the number of multimodal questions over time, particularly in the national subset. While early 2000s exams typically included only 2--4 visual items, recent exams, especially after 2017, often feature more than a dozen. This shift reflects evolving priorities in chemistry education, with greater emphasis on visual literacy, quantitative data interpretation, and apparatus reasoning as core components of interdisciplinary problem-solving.

This increase in visual content coincides with the recent performance surge of advanced MLLMs. Modern Olympiad questions, structured around diagrams and data-rich figures, align more closely with the strengths of models trained on large-scale image-text corpora. Human performance remains relatively stable across years, but the gap between humans and GPT-5 widens precisely in the more recent, visually intensive exams, underscoring how curriculum changes favor models with stronger multimodal integration.

At the model level, distinct behavioral patterns emerge. GPT-5 maintains consistently high accuracy across years, though occasional errors remain. GPT-4.1 is relatively stable, whereas GPT-4o exhibits pronounced year-to-year fluctuations, with both models often hovering near the estimated human average. These contrasts suggest that surpassing human performance is not yet universal across model families and highlight that robustness, rather than peak accuracy alone, is a critical dimension of progress.

Finally, because proprietary models use stochastic decoding, all per-year model accuracies are aggregated over multiple runs in our evaluation. This produces non-integer correct counts in Fig.~\ref{human-results} and provides more stable estimates across years. Combined with the temporal increase in multimodal content, a joint picture emerges: the Olympiad has shifted toward visually grounded scientific reasoning, and MLLMs show the greatest improvements in these areas. GPT-5 benefits most from this alignment, whereas GPT-4.1 and GPT-4o illustrate the remaining gap between achieving high peak scores and sustaining year-to-year stability. For chemists, the implications are twofold: Olympiad problems offer a sensitive probe of multimodal reasoning that reflects evolving chemistry education, and continued progress will require not only larger or newer models but also improved calibration and reliability for visually intensive, quantitatively precise tasks.

\begin{figure*}[ht]
  \centering
  \includegraphics[width=1\textwidth]{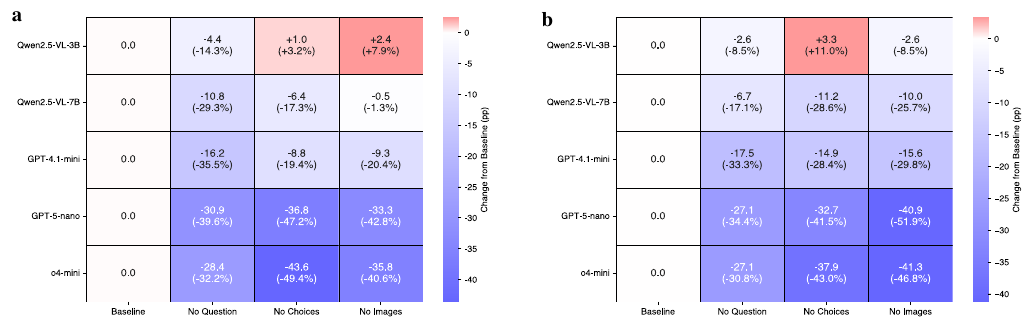}
  \caption{{\bf Ablation studies on local and national sets.} \footnotesize Each cell shows the performance drop from the baseline when one input content (question, choices, or image) is removed. The percentage change is shown alongside absolute differences. Blue color indicates degradation, and red indicates improvement. {\bf a,} Local set. {\bf b,} National set.}
    \label{fig-ablation}
\end{figure*}

\subsection*{When Seeing Hurts: Visual Input Can Degrade Performance}

To better understand the contribution of each input modality in USNCO-V, inspired by the practice in machine reading comprehension \cite{cui-etal-2022-iscience}, we perform an ablation study by systematically removing question text, answer choices, or image inputs during inference. Fig.~\ref{fig-ablation} visualizes performance changes for each model relative to their full-input baselines across the local and national sets. We select a representative set of models spanning different capability tiers: Qwen2.5-VL-3B and 7B as small open-source baselines, GPT-4.1-mini as a mid-capacity proprietary system, GPT-5-nano as a lightweight but reasoning-oriented proprietary model, and o4-mini as a high-performing proprietary baseline. This spectrum allows us to probe how modality dependence varies across model scale, architecture, and alignment.

Removing the question has a strong impact on mid-tier and top-tier models, reaffirming the foundational role of textual context in multimodal reasoning. Without a clear problem statement, models cannot anchor their decision-making process, leading to sharp declines. For example, GPT-5-nano drops by 30.9 pp on the local set and 27.1 pp on the national set, while o4-mini falls by 28.4 pp and 27.1 pp, respectively. Smaller models such as Qwen2.5-VL-3B and 7B show lighter declines, suggesting they rely less heavily on question text and more on surface-level heuristics.

Stripping away the answer choices leads to similarly large degradations, particularly in high-performing models. The o4-mini drops by 43.6 pp on the local set and 37.9 pp on the national set, while GPT-5-nano declines by 36.8 pp and 32.7 pp. GPT-4.1-mini also shows notable sensitivity (-8.8 to -14.9 pp). Interestingly, the smallest model, Qwen2.5-VL-3B, improves slightly when deprived of choices (+1.0 local, +3.3 national), indicating that poorly aligned models may be distracted by spurious choice formatting.

Removing images reveals a split between weak and strong models. In some cases, smaller systems improve slightly when the image is removed (e.g., Qwen2.5-VL-3B on the local set, +2.4 pp), suggesting that visual content introduces noise for models lacking robust alignment. By contrast, top-tier models degrade sharply: o4-mini drops by 35.8 pp (local) and 41.3 pp (national), and GPT-5-nano by 33.3 pp and 40.9 pp. GPT-4.1-mini exhibits a balanced profile, with moderate declines by 9.3 and 15.6 pp, consistent with its transitional capacity. Importantly, across models, the accuracy loss from image removal is consistently larger on the national set than on the local set, confirming that the national exams rely more heavily on visual reasoning, which matches our observations in baseline evaluations.

Viewed together, these results offer a tiered panorama of multimodal reasoning. Small models can sometimes gain when vision or choices are removed, reflecting distraction effects. Mid-tier models benefit moderately from all inputs but lack deep integration. Top models, by contrast, rely heavily on full context, especially images in the national subset, underscoring their dependence on advanced cross-modal fusion. This layered perspective suggests that one-size-fits-all strategies in model development are suboptimal.

In summary, these findings challenge the common assumption that visual input always enhances performance. Instead, they reveal that the utility of each modality depends on model capacity, fusion quality, and exam subset. Future MLLM development should prioritize robust modality alignment, architectural synergy, and task-aware prompting strategies to fully unlock multimodal reasoning capabilities across the capability spectrum.

\section*{Discussion}

\begin{figure*}[ht!]
  \centering
  \includegraphics[width=1\textwidth]{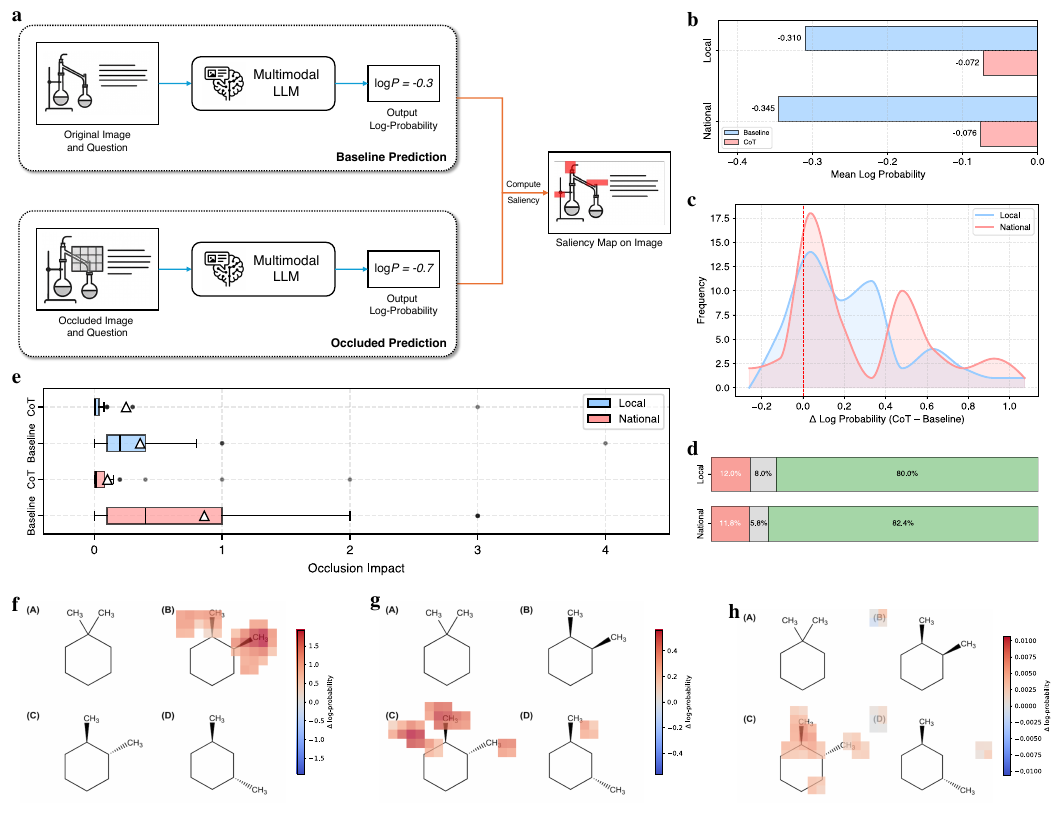}
  \caption{{\bf Occlusion-based saliency analysis on USNCO-V.}
  	\footnotesize
	{\bf a,} Pipeline of occlusion-based saliency analysis. The original image and question are fed to a multimodal LLM to obtain the chosen option and its log-probability. We then slide a $k\times k$ black occluder over the image and, at each location, measure the drop in log-probability for the chosen option (baseline - occluded). Aggregating these values over the grid yields a saliency map that highlights regions most influencing the model’s prediction.
	{\bf b,} Mean log-probabilities for baseline versus chain-of-thought reasoning on local and national subsets. A higher log-probability (less negative) indicates greater model confidence. CoT consistently yields higher confidence than the baseline model, showing the benefit of step-by-step reasoning.
	{\bf c,} Distribution of sample-level confidence gains, measured as $\Delta$ log-probability (CoT - Baseline). Positive values indicate that CoT reasoning increases model confidence. Most samples exhibit positive $\Delta$, with only a small fraction showing negative or negligible changes.
	{\bf d,} Win-rate comparison showing the proportion of samples where CoT reasoning outperforms the baseline, where the baseline is better, or where their performance is equivalent. Green represents CoT wins, red represents baseline wins, and gray indicates ties, based on a thresholded difference in log-probability ($\epsilon$ = 0.001).
	{\bf e,} Comparison of occlusion impact (log-probability change) distributions between Baseline and CoT models across local and national datasets. Each box summarizes the magnitude of prediction sensitivity when image regions are masked, with triangles marking the mean values and dots representing outliers. Blue corresponds to the local dataset and red to the national dataset.
	{\bf f-h,} Occlusion-based saliency maps for a stereochemistry question. Each panel shows the top-15 influential image regions based on the prediction change when occluded. The colorbar denotes $\Delta$ log-probability as a drop (baseline - occluded): red ($\Delta>0$) marks supportive regions (masking lowers confidence), blue ($\Delta<0$) marks suppressive regions (masking raises confidence). For clarity, overlays highlight the top-$N$ most impactful patches ranked by $|\Delta|$. In representative cases, the baseline concentrates on narrow local cues, whereas CoT exhibits broader cross-option attention with sharper localization of chemically relevant bonds and substituents and suppression of distractors.
	{\bf f,} Baseline visualization with its original prediction B.
  	{\bf g,} Baseline visualization when forcibly targeting correct option C.
  	{\bf h,} CoT visualization with its original correct prediction C.
  	}
  \label{fig-occlusion-overview}
\end{figure*}

Our comprehensive evaluation of multimodal large language models on chemistry Olympiad questions reveals both impressive capabilities and persistent limitations. Top proprietary models, such as GPT-5, consistently surpass the estimated performance of USNCO participants across multiple years, including on visually intricate, interdisciplinary chemistry problems. However, this success masks enduring challenges in integrating vision and language. Our ablation studies show that removing images can sometimes improve performance, suggesting that visual inputs are not always processed constructively and may introduce spurious cues.

Prompt engineering emerges as a critical strategy to address this issue. While few-shot prompting yields modest gains, CoT prompting consistently improves accuracy by scaffolding explicit reasoning. More importantly, CoT transforms the qualitative reasoning behavior of MLLMs, shifting them from pattern-matching heuristics to structured, comparative evaluation across options.

This transformation is evident in our confidence and interpretability analyses (Fig.~\ref{fig-occlusion-overview}). 
To ensure a fair comparison, we systematically analyze only the samples that are simultaneously predicted correctly by both the baseline and CoT models, allowing us to better capture their behavior under the same conditions.
Although both settings yield correct predictions, CoT consistently shows higher model confidence, where mean log-probabilities improve from -0.310 to -0.072 on the local set and from -0.345 to -0.076 on the national set (Fig.~\ref{fig-occlusion-overview}b). 
The distribution of per-sample confidence gains (Fig.~\ref{fig-occlusion-overview}c) shows that the majority of samples achieve improvements (median $\approx$0.15). Both subsets show right-skewed distributions, but with distinct allocation of mass: the local set concentrates in moderate gains ($\approx$0.11--0.41; 40\%), whereas the national set is more polarized, with more near-zero cases (35\%) and a much heavier large-gain tail ($\ge$0.41; 39\%), pushing the upper quartile from 0.35 (local) to 0.50 (national). This indicates that CoT delivers steady benefits on typical items and very large boosts on the most challenging national questions.
Complementing these views, win-rate analysis confirms CoT’s broad reliability: approximately 80--82\% of samples favor CoT in both datasets, while only about 12\% favor the baseline and $\le$8\% are ties (Fig.~\ref{fig-occlusion-overview}d).

To illustrate what these gains look like qualitatively, we show a representative stereochemistry problem (Fig.~\ref{fig-occlusion-overview}f-h).
In the baseline model, attention is narrowly confined to a single superficial cue, such as one methyl group, while neglecting stereogenic centers and their spatial orientation. This over-reliance on a narrow set of visual features reflects brittle pattern matching. Even when the model is nudged to consider the correct stereochemical configuration (option C), its focus broadens only slightly with minor spillover to option D, and remains constrained, indicating shallow, non-comparative processing.
With CoT prompting, attention becomes more discriminative and distributed, highlighting stereochemically relevant features such as multiple substituents around the stereogenic center, while simultaneously suppressing irrelevant distractors. This broader focus enables the model to integrate the relative spatial arrangement of groups, mirroring expert elimination strategies in stereochemical reasoning.

This visual shift mirrors the broader quantitative findings shown in Fig.~\ref{fig-occlusion-overview}e.
Baseline models exhibit large, sharply peaked occlusion impacts, indicating narrow focus and brittle reliance on a few dominant cues. In contrast, CoT models distribute their reliance more evenly across multiple relevant regions, particularly in the national set, which demands higher-order reasoning. Although baseline heatmaps may appear visually “cleaner,” this narrow focus overlooks key stereochemical determinants. CoT’s more diffuse saliency maps instead capture integrative reasoning, comparing spatial evidence across options in a way that reflects expert comparative evaluation.
Thus, higher performance with CoT does not result from simply sharpening attention but from fundamentally restructuring how visual evidence is processed through a more integrated, comparative reasoning framework.

When considered alongside our ablation findings, this leads to a central insight: visual inputs alone are insufficient for robust multimodal reasoning. Without structured guidance, MLLMs may underutilize or misinterpret images. CoT prompting enables visual input to function as a reliable asset, supporting the comparative reasoning required for authentic chemical analysis. These observations underscore a broader design principle: model architectures and training should incorporate reasoning-aware objectives that align modalities through interpretable chains of logic.

This insight resonates with trends in STEM education. Over the past two decades, the USNCO exams have increasingly incorporated multimodal elements, particularly in the national set. CoT-enhanced MLLMs perform well under these evolving demands, suggesting potential not only as solvers but as intelligent teaching assistants capable of decomposing complex, real-world scientific tasks.

Nonetheless, challenges remain. Even with CoT, our saliency maps reveal occasional failures, such as missing one of the key substituents in stereochemical diagrams or focusing on irrelevant components. This suggests that more substantial domain alignment is needed, potentially through fine-tuning chemistry-specific tasks, integrating symbolic constraints, or making architectural improvements in cross-modal fusion.

In parallel, retrieval-augmented generation (RAG) represents another promising direction. By integrating external chemistry resources such as textbooks, laboratory manuals, or past Olympiad solutions, RAG could provide complementary knowledge to support reasoning on complex multimodal tasks. While our current study isolates intrinsic model capability without external context, USNCO-V offers a natural testbed for future work exploring how RAG can enhance multimodal problem solving.

Looking forward, we envision extending CoT frameworks to multi-step problem chains, incorporating symbolic chemistry engines for mechanistic validation, and embedding spatial reasoning modules for better diagram interpretation. By coupling interpretability methods such as occlusion analysis with model training and design, future systems can become not only more accurate but also transparent, trustworthy, and aligned with educational priorities.

While USNCO-V provides a rigorous and diverse testbed for evaluating multimodal reasoning, it represents one specific context within the broader landscape of chemistry education and research. 
Olympiad-style questions emphasize conceptual integration, diagram interpretation, and symbolic reasoning, which make them ideal for probing visual-textual alignment in large language models but do not encompass all forms of chemical inquiry, such as laboratory practice, materials discovery, or data-driven prediction. The present study therefore captures a distinct yet important facet of multimodal chemistry evaluation. 
Future work could explore complementary benchmarks, such as other national and international Olympiads (e.g., IChO, UKChO, Australian Chemistry Olympiad) and curricular assessments like AP Chemistry and GRE Chemistry, to investigate how multimodal reasoning generalizes across educational levels and problem types.

\small
\section*{Methods}

\subsection*{Dataset Curation}

We constructed USNCO-V by manually collecting multimodal chemistry questions from publicly available U.S. National Chemistry Olympiad (USNCO) exams. Our source material spans local exams from 2000 to 2025 and national exams from 1999 to 2025, each comprising 60 multiple-choice questions in Part~I. We focus exclusively on these multiple-choice components due to their standardized structure and objective scoring.

Because exam PDFs vary greatly in layout and formatting, automated extraction tools such as \texttt{pymupdf} proved unreliable. To preserve content fidelity, all question texts and associated images were manually curated. Text was transcribed directly, and images were captured using high-resolution screenshots. This process yielded a total of 204 local and 269 national questions, each paired with one image and four answer choices.

For fine-grained analysis, we further annotated each image with one of five visual categories: table, chart, molecular structure, apparatus, and other. These labels support the task-type performance breakdowns presented in the main paper.

Table~\ref{data-stats} summarizes the dataset statistics. National questions tend to have slightly longer text on average and marginally smaller image dimensions, likely reflecting formatting constraints. All images are retained in their original resolution, with resizing deferred to model-specific preprocessing.

We also examined distributional properties such as image type frequency and answer key balance (Fig.~\ref{usnco-v-overview}b and Fig.~\ref{usnco-v-overview}c). The distribution of correct options is uniform across A--D choices, mitigating label bias. Visual modality varies by exam type: national exams include a higher proportion of data-driven charts and analytical figures, while local exams show relatively more straightforward molecular structures.

\begin{table}[h]
\small
\caption{\label{data-stats} {Statistics of the curated USNCO-V dataset.}}
\begin{center}
\begin{tabular}{l c c }
\toprule
{\bf Statistics} & {\bf Local} & {\bf National}   \\ 
\midrule
Year					& 2000--2025 & 1999--2025 \\
\midrule
Total Questions			& 204 & 269 \\
Question Tokens (avg) 	& 35.2 & 38.7 \\
Question Tokens (max) 	& 141 & 139 \\
\midrule
Number of Choices		& 4 & 4 \\
Choice Tokens (avg)		& 31.3 & 31.2 \\
Choice Tokens (max) 		& 133 & 120 \\
\midrule
Image Size (avg, px)		& 1234$\times$630  & 1077$\times$663 \\
Image Size (max, px)		& 2809$\times$2328 & 2351$\times$2530 \\
\bottomrule
\end{tabular}
\end{center}
\end{table}

\subsection*{Experimental Setups for Baselines}

We benchmarked a diverse set of state-of-the-art multimodal large language models (MLLMs), covering both proprietary and open-source systems. Proprietary models include the GPT series (GPT-5\cite{gpt-5}, o3\cite{o3-o4mini}, o4-mini\cite{o3-o4mini}, o1\cite{jaech2024openai}, GPT-4.1\cite{gpt41}, GPT-4o\cite{hurst2024gpt}), and Gemini series (Gemini-2.5-Pro\cite{google2025gemini25pro}, Gemini-2.5-Flash\cite{google2025gemini25pro}). 
All proprietary models were accessed through their official APIs (OpenAI and Google Gemini). When stochastic decoding was unavoidable (due to non-deterministic API defaults), we repeated inference five times to support pass@5 and consistency@5 evaluation.

Open-source models include Qwen2.5-VL\cite{bai2025qwen25vl} (sometimes we omit ``-Instruct'' suffix for simplicity), InternVL3.5\cite{internvl35}, Intern-S1\cite{bai2025interns1}, MiniCPM-V 4.5\cite{minicpm-v-4}, GLM-4.1V\cite{hong2025glm41v}, Ovis2.5\cite{lu2025ovis2}, Molmo\cite{Deitke2024MolmoAP}, Phi-4-Multimodal\cite{phi4v}, Janus-Pro\cite{chen2025janus}, and Gemma-3\cite{gemma3}. For chemistry-specific models, we additionally benchmarked ChemVLM\cite{li2025chemvlm} (8B and 26B). All open-source models were run locally using the Hugging Face \texttt{transformers} library on NVIDIA A100 or H100 GPUs without quantization. We used a batch size of 1 for evaluation and greedy decoding (temperature = 0) to ensure deterministic predictions.

All models were evaluated using the original question images without modification. Any preprocessing, such as resizing or normalization, was handled internally by the model’s vision encoder. The text inputs included the question stem and four answer choices, formatted uniformly in the prompt. We used each model’s default image-text fusion and tokenization pipeline. 

For zero-shot evaluations, we employed a unified instruction template:

\begin{tcolorbox}[colback=gray!10, colframe=black, sharp corners]\small
Question: \{question\} \\
\{choices\} \\
* Answer with the option's letter (A, B, C, D) from the given choices directly without rationale.
\end{tcolorbox}

Most models follow the instruction and return a single option letter (A--D).
When extra text is present, we apply a strict regular expression to extract a standalone option letter.
If no valid option can be extracted, the response is treated as incorrect.

\subsection*{Evaluation Metrics and Significance Tests}

\textbf{Primary metric (accuracy).} 
We use accuracy as the primary metric, defined as the proportion of correctly predicted answers over the total number of questions. For a dataset with $N$ questions, let $\hat{y}_i$ denote the model's predicted option for question $i$ and $y_i$ the ground truth answer. 
Predictions outside the valid set \{A, B, C, D\} are treated as incorrect. 
The accuracy is:
\begin{equation}
	\text{Accuracy} \;=\; \frac{1}{N} \sum_{i=1}^{N} \mathbb{I}[\hat{y}_i = y_i],
\end{equation}
where $\mathbb{I}[\cdot]$ is the indicator function. 
This ensures that invalid outputs are penalized rather than ignored. 
We evaluate consistently across zero-shot, few-shot, and CoT prompting, and report results for the local and national subsets as well as their macro-average.

\textbf{Confidence intervals.} 
For each reported accuracy, we provide a 95\% confidence interval (CI) computed with the Wilson score interval. 
Let $s$ be the number of correct predictions among $n{=}N$ total questions and $\hat{p}=s/n$. 
With $z{=}1.96$,
\begin{equation}
\text{CI}_{95\%} = 
\frac{\hat{p} + \tfrac{z^2}{2n} \;\pm\; z \sqrt{\tfrac{\hat{p}(1-\hat{p})}{n} + \tfrac{z^2}{4n^2}}}{1+\tfrac{z^2}{n}}.
\end{equation}
We use this CI for all models. For deterministic runs (open-source models; temperature $=0$), this is the only uncertainty reported.

\textbf{Additional reliability metrics.} 
For proprietary models where multiple independent runs are possible, we also report pass@5 and consistency@5. 
For each question $i$, five independent predictions $\{\hat{y}_{i,1}, \ldots, \hat{y}_{i,5}\}$ are generated. 
Predictions outside the valid set \{A, B, C, D\} are treated as incorrect. 
The pass@5 score is defined as:
\begin{equation}
\mathrm{pass@5} = \frac{1}{N} \sum_{i=1}^{N} \mathbb{I}\!\left[\exists j:\hat{y}_{i,j} = y_i\right],
\end{equation}
where $N$ is the total number of questions. 
Questions for which all five outputs are invalid or incorrect are penalized.

For consistency@5, we compute for each question $i$ the maximum agreement among the five predictions:
\begin{equation}
m_i = \max_{a \in \{A,B,C,D\}} \frac{1}{5}\sum_{j=1}^{5} \mathbb{I}[\hat{y}_{i,j} = a],
\end{equation}
and then average across all questions:
\begin{equation}
\mathrm{consistency@5} = \frac{1}{N} \sum_{i=1}^{N} m_i.
\end{equation}
Invalid outputs do not match any valid option and thus reduce the agreement score for that question.
For open-source models, which are evaluated with deterministic decoding, we report only accuracy together with its 95\% confidence interval, as pass@5 and consistency@5 are not applicable.

\textbf{Significance testing.} 
In experiments where sampling introduces stochasticity (e.g., few-shot or CoT prompting), we conduct five independent trials using different sampled demonstrations. For these cases, we report the mean accuracy along with the standard error of the mean (SEM). Statistical significance is assessed using one-sided Welch's $t$-test \cite{welch1947generalization}, which is appropriate when comparing means with unequal variances. Significance levels are denoted as: $^*p{<}0.05$, $^{**}p{<}0.01$, $^{***}p{<}0.001$, $^{****}p{<}0.0001$.

\subsection*{Few-shot prompting}
We evaluate the effect of in-context learning through few-shot prompting, where $k$-shot exemplars ($k \in \{1,2,3,4,5\}$) are formatted as multi-turn dialogues. To avoid overlap between training and evaluation, exemplars are drawn from the opposite subset of the target set (i.e., national questions are used when evaluating on the local set, and vice versa). This design maintains task alignment while ensuring distributional separation.

Exemplars are selected using a sliding window strategy to balance diversity and reproducibility. For each $k$-shot setting, five independent runs are performed. In the 2-shot condition, for example, run~1 uses examples 0 and 1, run~2 uses examples 1 and 2, continuing until run~5, which uses examples 4 and 5. This procedure provides systematic coverage of candidate examples while reducing sensitivity to exemplar choice.

During inference, the model’s output is parsed to extract the predicted option letter and compared against the ground truth. For each $k$-shot configuration, we report the mean accuracy together with the standard error of the mean (SEM) across five runs. Statistical significance relative to the zero-shot baseline is assessed using one-sided Welch’s $t$-tests.

\subsection*{Chain-of-Thought prompting}
Chain-of-Thought (CoT) prompting elicits intermediate reasoning steps by appending a brief instruction, similar to “Let’s think step by step”, to the end of the question prompt. This encourages the model to produce a structured rationale before the final answer and can improve performance on tasks requiring logical reasoning \cite{kojima2022large}.
We apply CoT prompting to all evaluated models using the same base template as in the zero-shot setting, with a single reasoning trigger appended after the question text. An example is shown below:
\begin{tcolorbox}[colback=gray!10, colframe=black, sharp corners]
\small
Question: \{question\} \\
\{choices\} \\
* Firstly, analyze and solve the question step by step with rationale. \\
* Finally, answer with the option's letter (A, B, C, D) from the given choices directly without rationale.
\end{tcolorbox}

To assess the effect of CoT prompting, we compare accuracy with and without CoT on both the local and national subsets. 
For models with randomized decoding, each configuration is run five times. We report mean accuracy and the standard error of the mean (SEM). 
Statistical significance is evaluated using one-sided Welch's $t$-tests, consistent with our evaluation protocol.

Detailed performance and interpretability analyses of CoT prompting are provided in the corresponding results sections.

\subsection*{Occlusion-Based Interpretability}

To investigate how models utilize visual information during multimodal reasoning, we employ an occlusion-based saliency analysis \cite{zeiler2014visualizing,chefer2021transformer}. The method estimates the importance of an image region by measuring the change in the model’s log-confidence when that region is masked.

Let $I$ be the input image, $Q$ the accompanying question (including answer choices), and $f(I, Q)$ the predicted distribution over $\mathcal{A}=\{A, B, C, D\}$. For each item, we target the chosen option $a^*$ (in jointly-correct analyses, $a^*$ equals the ground truth answer). We define a square occlusion mask $M_{(i, j)}$ of size $k\times k$ pixels (centered at $(i, j)$) and construct the occluded image
\begin{equation}
\tilde{I}_{(i,j)} \;=\; I \odot \bigl(1 - M_{(i,j)}\bigr) \;+\; \mathbf{0}\cdot M_{(i,j)},
\end{equation}
i.e., a black fill (\(0,0,0\)) is used for all computations. We then measure the log-probability drop
\begin{equation}
S(i,j) \;=\; \log f_{a^*}(I,Q) \;-\; \log f_{a^*}\!\bigl(\tilde{I}_{(i,j)}, Q\bigr)\!,
\end{equation}
so that $S(i,j){>}0$ indicates a supportive region (occluding it reduces confidence in $a^*$), while $S(i,j){<}0$ indicates a suppressive region. Saliency maps are computed on a sliding grid with patch size $k{=}56$ and stride $s{=}28$.

For visualization, we render a zero-centered, symmetric color scale (diverging colormap) as an alpha-blended heatmap overlay on the original image. To improve readability, the overlay highlights the top-$N$ most impactful patches, ranked by the magnitude of the change in log-probability, $|S(i,j)|$; red ($S>0$) denotes supportive regions (masking reduces confidence), whereas blue ($S<0$) denotes suppressive regions (masking increases confidence). As a scalar summary per question, we report the occlusion impact $r=\max_{(i,j)}|S(i,j)|$ and aggregate $\{r\}$ across items (Fig.~8e). The procedure is model-agnostic and requires only black-box access to option log-probabilities or logits.

\section*{Data availability}

The original exam PDFs are publicly available through the American Chemical Society archive:
\url{https://www.acs.org/education/olympiad/prepare-for-exams.html}.
We provide curated metadata for the USNCO-V benchmark, including year, exam type, original question number, answer key, image category, and a brief question summary.
The complete metadata collection is archived on Zenodo (\url{https://doi.org/10.5281/zenodo.17139984}).
For reproducibility, the Zenodo repository includes detailed guidelines for manually extracting diagrams from publicly available exam PDFs.
Due to copyright restrictions, no original exam images or questions are redistributed.
All figures in this paper are illustrative mock-ups created solely for demonstration purposes and distinct from the actual benchmark materials.
Source data of charts are provided in Supplementary Data 1.

\section*{Code availability}

The source code is publicly available on Zenodo (\url{https://doi.org/10.5281/zenodo.17139984}).
The repository includes evaluation scripts that operate on complete question-image inputs.
Because of copyright restrictions, only metadata is provided. 
Users can reproduce results by obtaining the original exam materials from the ACS archive and following the data-processing instructions detailed in our repository.

\small
\bibliography{new.bib}

\section*{Acknowledgements}

This work is supported by the National Key Research and Development Program of China (2024YFC3308202).
X.L. acknowledges the support from the Anhui Province Science and Technology Innovation Project (202423k09020010).

\section*{Author contributions}

Conceptualization and methodology were led by Y.C.
Dataset curation and examination were performed by Y.C. and Y.Q.
Investigation and analysis were carried out by Y.C. and X.Y.
The original draft was written by Y.C. with input from all authors.
Review and editing were conducted by Y.C., X.Y., Y.Q., X.L., S.W., and G.H.
Funding acquisition and supervision were provided by S.W. and G.H.

\section*{Competing interests}
The authors declare no competing interests.

\end{document}